\documentclass[sigconf]{acmart}

\usepackage{multirow}
\usepackage{colortbl}
\usepackage{xcolor}
\usepackage{graphicx}
\definecolor{c1}{HTML}{ffcc99}
\definecolor{c2}{HTML}{fff8ae}

\newcommand{\figref}[1]{Fig.~\ref{#1}}
\newcommand{\tabref}[1]{Table~\ref{#1}}

\AtBeginDocument{%
  }

\setcopyright{acmlicensed}
\copyrightyear{2024}
\acmYear{2024}
\setcopyright{acmlicensed}\acmConference[SIGGRAPH Conference Papers '24]{Special Interest Group on Computer Graphics and Interactive Techniques Conference Conference Papers '24}{July 27-August 1, 2024}{Denver, CO, USA}
\acmBooktitle{Special Interest Group on Computer Graphics and Interactive Techniques Conference Conference Papers '24 (SIGGRAPH Conference Papers '24), July 27-August 1, 2024, Denver, CO, USA}
\acmDOI{10.1145/3641519.3657456}
\acmISBN{979-8-4007-0525-0/24/07}

\begin{document}

\title{3D Gaussian Splatting with Deferred Reflection}

\author{Keyang Ye}
\orcid{0009-0005-8675-566X}
\affiliation{%
  \institution{State Key Lab of CAD\&CG\\Zhejiang University}
  \city{Hangzhou}
  \country{China}
}
\email{yekeyang@zju.edu.cn}

\author{Qiming Hou}
\orcid{0009-0004-4177-4704}
\affiliation{%
  \institution{State Key Lab of CAD\&CG\\Zhejiang University}
  \city{Hangzhou}
  \country{China}
}
\email{hqm03ster@gmail.com}

\author{Kun Zhou}
\authornote{Corresponding author}
\orcid{0000-0003-4243-6112}
\affiliation{%
  \institution{State Key Lab of CAD\&CG\\Zhejiang University}
  \city{Hangzhou}
  \country{China}
}
\email{kunzhou@acm.org}

\begin{abstract}
  The advent of neural and Gaussian-based radiance field methods have achieved great success in the field of novel view synthesis. However, specular reflection remains non-trivial, as the high frequency radiance field is notoriously difficult to fit stably and accurately. We present a deferred shading method to effectively render specular reflection with Gaussian splatting. The key challenge comes from the environment map reflection model, which requires accurate surface normal while simultaneously bottlenecks normal estimation with discontinuous gradients. We leverage the per-pixel reflection gradients generated by deferred shading to bridge the optimization process of neighboring Gaussians, allowing nearly correct normal estimations to gradually propagate and eventually spread over all reflective objects. Our method significantly outperforms state-of-the-art techniques and concurrent work in synthesizing high-quality specular reflection effects, demonstrating a consistent improvement of peak signal-to-noise ratio (PSNR) for both synthetic and real-world scenes, while running at a frame rate almost identical to vanilla Gaussian splatting.
\end{abstract}

\ccsdesc[500]{Computing methodologies~Rasterization}
\ccsdesc[500]{Computing methodologies~Point-based models}
\ccsdesc[500]{Computing methodologies~Machine learning approaches}
\ccsdesc[300]{Computing methodologies~Rendering}

\keywords{Novel view synthesis, deferred shading, real-time rendering}

\begin{teaserfigure}
  \includegraphics[width=1.0\columnwidth]{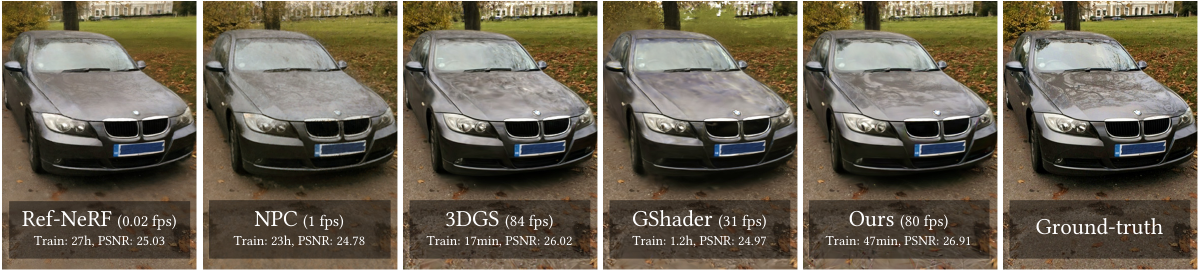}
  \caption{Novel view synthesis with specular reflection. From left to right: Ref-NeRF~\cite{ref_nerf}, Neural Point Catacaustics~\cite{catacaustics}, 3D Gaussian Splatting~\cite{3DGS}, GaussianShader~\cite{jiang2023gaussianshader}, our method, ground-truth.
  We render high quality reflection at a speed comparable with the original reflection-oblivious 3D Gaussian Splatting. The key contribution is a \emph{deferred shading} pipeline, which offers high-precision shading in real-time and enables gradual propagation of correct normal estimation. }
  \Description{This figure presents novel view synthesis results of a real scene, produced by different methods. Our method renders high quality reflection at 80 frames per second, comparable with the original 3D Gaussian Splatting method.}
  \label{fig:teaser}
\end{teaserfigure}

\maketitle

\section{Introduction}

Novel view synthesis of 3D scenes captured with multiple images has been a long-standing research topic in computer graphics and vision. Recently, Neural Radiance Field (NeRF) methods first introduced by Mildenhall et al.~\shortcite{nerf} have gained popularity by using volume rendering with implicit fields via Multi Layer Perceptrons (MLPs), achieving state-of-the-art visual quality. More recently, 3D Gaussian Splatting \cite{3DGS} (3DGS) offers a more compelling solution by modeling radiance fields as sparsely distributed 3D Gaussians, synthesizing novel views at high resolution and real-time frame rates, while maintaining state-of-the-art visual quality and competitive training.

However, specular reflection remains challenging for Gaussian splatting to model. Although 3DGS provides view-dependent coloring via per-Gaussian SH (Spherical Harmonics) functions, its directional frequency is too limited to model specular reflection. The training process instead hallucinates Gaussians to explicitly model the reflected image, which lacks a well-defined spatial position for non-planar reflectors. As such, specular effects end up poorly emulated, with a side effect of compromising geometry quality.

In this paper, we introduce a deferred shading method to effectively render specular reflection with Gaussian splatting. Our method associates each Gaussian with a scalar parameter of reflection strength and regards the shortest axis of each Gaussian ellipsoid as its normal vector. The rendering is performed in two passes. First, a Gaussian splatting pass generates several screen-space maps of base color, normal, and reflection strength. Second, a pixel shading pass queries an environment map with the reflection direction to acquire the specular reflection color, and renders the final color as the sum of the basic and reflection colors weighted by the reflection strength. The environment map, per-Gaussian reflection strength, as well as other Gaussian parameters are all learned during training.

The seemingly mundane environment map query presents significant challenge to the training procedure. As a high frequency lookup table, the environment map places a high precision demand on the normal vectors required to compute reflection directions, while scarcely providing any useful gradient to refine them. On top of that, we only have a semi-transparent Gaussian soup with loosely-defined surface. To this end, we present a training algorithm featuring \textit{normal propagation}. Specifically, based on the observation that Gaussians with relative large reflective strength values have near-correct normal vectors, we expand these reflective Gaussians to propagate their normal vectors to nearby Gaussians. In this way, after one Gaussian with near-correct normal overlaps a different Gaussian without one, some shared pixels can also have near-correct normal, which will get meaningful normal gradients, helping to optimize the normal of the later Gaussian.

Our deferred shading model is critical to the efficacy of training. The Gaussian splatting pass blends Gaussian properties like base color and normal into viewport-aligned textures. The blended input values on each pixel are used to evaluate reflection and compose the base and reflection colors into the final color, which feeds gradient back to input values of the same pixel through image color loss. This creates a gradient channel from color to blended normal to individual Gaussian normal, facilitating information flow between the normal of different Gaussians overlapping the same pixel, which enables normal propagation. This is not possible with per-Gaussian shading, where gradients propagate from color to individual Gaussian normal directly and different Gaussians get independent normal gradients that cannot influence each other.

Experimental results on several previously published datasets show that our method significantly outperforms state-of-the-art methods and concurrent work in synthesizing high-quality specular reflection effects (see \figref{fig:teaser}), demonstrating a consistent improvement of peak signal-to-noise ratio (PSNR) for both synthetic and real-world scenes, while running at real-time frame rates almost identical
to vanilla Gaussian splatting. 
Our method also produces more accurate normal and environment map estimation (see \figref{fig:normal} and \figref{fig:envmap}). On the other hand, it does not provide a full geometry-lighting-material decomposition for inverse rendering or relighting. Our shading model separately handles mirror reflection and leaves rough reflection, anisotropic / layered materials, and global illumination effects to the base SH colors, achieving high-quality rendering superior to full-decomposition methods in novel view synthesis.

\section{Related work}

Research on novel view synthesis has a long history in computer graphics and vision, and is still developing rapidly. In this section we only review the most relevant references. Please refer to \cite{NeuralField-survey-CGF2022,NeuralRendering-survey-CGF2022} for comprehensive reviews of the field.

\paragraph{Novel View Synthesis.} A variety of methods have been proposed to synthesize novel views from multiple images of a static scene~\cite{lightfield1996,lumigraph1996,davis2012unstructured}. Recently, Neural Radiance Field (NeRF)~\cite{nerf} has gained popularity. NeRF represents the scene as implicit fields of view-dependent color and density, typically evaluated as a deep MLP, and synthesizes high-quality images through volumetric ray marching.

Many follow-up research on NeRF try to improve the image quality (e.g., \cite{mip_nerf_360,ref_nerf}), training/rendering performance (e.g., \cite{instant_ngp, yu2021plenoctrees, fast_nerf}), sparse-view generalization ability (e.g., \cite{Deng2022cvpr,Jain2021cvpr,Chen2021cvpr}).
One notable work~\cite{instant_ngp} replaces the deep MLP of classical NeRF with a shallow MLP featuring multi-resolution hash encoding as input, which enables rapid training and real-time rendering. As a generic neural network design with applications beyond novel view synthesis, it is orthogonal to our approach, which does not use neural networks.

Rendering novel views of reflective objects is challenging as the radiance field becomes high frequency in the view dimension, making it difficult to reconstruct from the spatial images typically used as input. Classical NeRFs assume low frequency view-dependency. Ref-NeRF~\cite{ref_nerf} extends NeRFs with a novel parameterization for view-dependent outgoing radiance and normal vector regularization. The added dimensions are challenging to optimize, though, which makes its training time-consuming and leads to noisy synthesis results.

The radiance field can also be represented by a point cloud~\cite{xu2022point,Zhang2022sa}, which enables unique algorithm designs. Kopanas et al.~\shortcite{catacaustics} formalize hallucinated mirror images as catacaustics consisting of virtual points inside the reflector. Curved surfaces are handled using MLPs to adjust the point positions based on camera movement. However, the training of this MLP is typically under-constrained due to static input cameras, resulting in unstable behavior on novel views.

The more recent 3D Gaussian splatting~\cite{3DGS} departs from the volumetric NeRF formulation in favor of a differentiable rasterizer for spatial Gaussians, achieving a distinctive hard-realtime frame rate at 1080p resolutions. Our approach extends its Gaussian representation, significantly enhancing specular reflection effects while maintaining the original training and rendering efficiency.

\begin{figure*}
\includegraphics[width=1.0\linewidth]{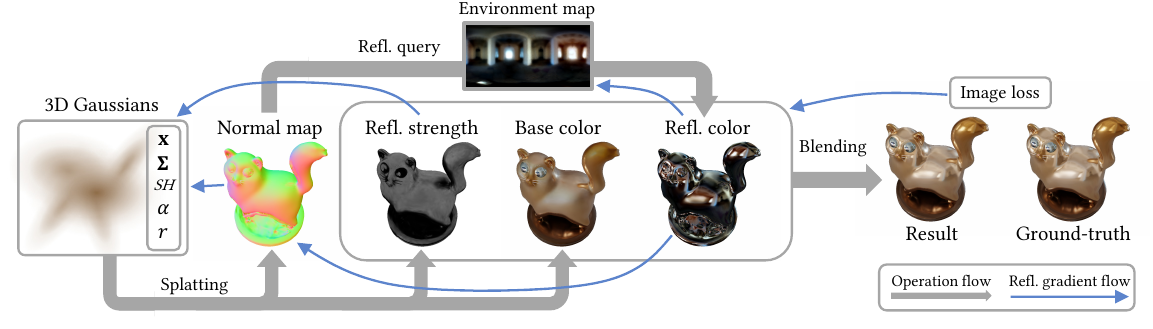}
\caption{Our rendering pipeline. A Gaussian splatting pass is first performed to bake reflection strength, normal, and base color to screen space maps. In the following shading pass, for each pixel, we use the normal map to compute a reflection direction and query an environment map for a reflected color. The reflection strength is then used to blend base color and reflection color into the final result. An image loss is used to back-propagate gradients. Note that there exist many gradient propagation paths. Here we only illustrate the gradient flow most relevant to reflection fitting.\label{fig:components}}
\Description{On the far left of the figure, symbols representing different Gaussian properties are listed, along with a schematic of the Gaussians. Arrows point from the Gaussian schematic to the splatted-out normal, reflection strength, and base color maps. Another arrow leads from the normal map through the environment lighting map to the reflection color map. The reflection strength, base color, and reflection color maps are enclosed, with arrows pointing to the final result image labeled ``Blending''. On the right side of the result image is the ground truth, with ``Image loss'' written between them. Gradient flows are depicted with arrows from the image loss to the reflection color map, from the reflection color map to the environment lighting map and normal map, and from the normal map and reflection strength map to the Gaussian schematic.
}
\end{figure*}

\paragraph{Inverse Rendering.}  It is tempting to generalize novel view synthesis to \emph{inverse rendering}~\cite{nerfactor, nerv, nerd, physg, zhang2022modeling, jiang2023gaussianshader}, 
which aims to completely decompose geometry, lighting and material, typically employing a physically-based shading model.
While such a scene model can automatically render a wide range of visual effects including reflection, its optimization is significantly under-constrained, especially when starting from a few static images. The corresponding optimization tends to rely on prior assumptions about the scene, such as material or geometry smoothness, compromising the overall quality when applied to novel view synthesis.

Recent Signed Distance Field (SDF) methods~\cite{envidr, nvidiffrec, nero} utilize the NeRF neural field as a geometry representation, leading to an elegant inverse rendering formulation that allows flexible lighting and material changes. The flexibility comes at a cost, as they inherit the high training cost of NeRF and the inherent smoothness of neural SDFs tends to over-smooth geometry details.

The concurrent work of GaussianShader~\cite{jiang2023gaussianshader} approaches the inverse rendering goal using a Gaussian-based scene representation, estimating physically-based BRDF parameters and a normal vector for each Gaussian. Their method inherits the real-time performance of 3DGS, albeit with noticeable overhead.

Both our method and GaussianShader tackle the same specular reflection problem in a Gaussian splatting setting. The key difference is that we compute reflection on pixels as opposed to Gaussians, generating more reflection samples for the same rendering cost. The extra samples stabilize reflection strength and environment map training by smoothing out gradient, leading to significantly less noise in the final result. Our per-pixel shading also eliminates interpolation artifacts caused by discontinuous color change at Gaussian boundaries.

\paragraph{Deferred Shading.} Deferred shading~\cite{DeferredRendering} is a classical real-time rendering technique that computes high frequency shading effects like specularity per-pixel in screen space, after baking scene properties like positions and normal into viewport-aligned textures. Representing a scene as a neural texture map on top of 3D meshes, the deferred neural rendering approach~\cite{DeferredNeuralRendering2019} first rasterizes the scene into a screen-space feature map, which is then converted to photo-realistic images based on a neural network. Both the neural network and the neural texture are trained end-to-end. Hedman et al.~\shortcite{DeferredNerf2021} propose a deferred NeRF architecture, which renders screen-space diffuse color and feature vector maps using a deep MLP, followed by another shallow MLP to predict the view-dependent specular color for each pixel. 
We combine deferred shading with 3D Gaussian splatting to effectively render specular reflection, which also produces accurate estimation of normal and environment maps.

\section{Method}

\subsection{Rendering Model}

Our deferred rendering model consists of two passes. The first is Gaussian splatting.
Following the original setting of the 3DGS renderer~\cite{3DGS}, we start with Gaussian parameters ${\Theta}_i$, per-Gaussian view-dependent SH colors ${c}_i(\mathbf{v})$, and compute the pixel colors ${C}(\mathbf{v})$. Here $i$ is the Gaussian index and $\mathbf{v}$ refers to the view direction. For simplicity, we also parameterize the output image with $\mathbf{v}$, and treat the splatting process as a blackbox that blends colors with linear weights $G$:

\begin{equation}
C(\mathbf{v})=\sum_i{{c}_i(\mathbf{v}) G({\Theta}_i,\mathbf{v})}.
\label{equ:splat}
\end{equation}

Here the most expensive component is the weights $G$. It is computed at the finest per-Gaussian-per-pixel granularity, is order-sensitive and requires a sort as a preprocess. With $G$ already computed, though, it is very cheap to blend extra per-Gaussian values alongside $c_i$. We apply this to $n_i$, the shortest axis of each Gaussian ellipsoid interpreted as its normal vector, and $r_i$, a per-Gaussian scalar controlling its specular reflection strength:
\begin{equation}
N(\mathbf{v})=\sum_i{n_i G({\Theta}_i,\mathbf{v})}, ~~R(\mathbf{v})=\sum_i{r_i G({\Theta}_i,\mathbf{v})},
\label{equ:normal}
\end{equation}
where $n_i$ are flipped as necessary to face the camera.

Second, a deferred reflection pass composes the final pixel color $C'(\mathbf{v})$, detached from Gaussians $G$:
\begin{equation}
C'(\mathbf{v})=(1-R(\mathbf{v}))C(\mathbf{v})+R(\mathbf{v})E(\frac{2\mathbf{v} \cdot N(\mathbf{v}) N(\mathbf{v})}{||N(\mathbf{v})||}-\mathbf{v}),
\label{equ:deferred}
\end{equation}
where $E$ is a learned environment map queried on the reflection direction with a bilinear filter.

\figref{fig:components} illustrates the components used by our rendering model. The per-Gaussian normal $n_i$ and reflection strength $r_i$ are trained and splatted separately. The splatted images are combined in the shading pass to compose the final image. This process works entirely in screen space. The environment map is trained entirely from the final pass, detached from the Gaussians.

\subsection{Loss Function and Normal Gradient}

When training, we use the same combined image loss $\mathcal{L}_1$ and D-SSIM loss functions as in \cite{3DGS}:
\begin{equation}
\mathcal{L} = (1-\lambda)\mathcal{L}_1 + \lambda\mathcal{L}_{D-SSIM},
\label{equ:loss}
\end{equation}
where $\lambda = 0.2$ in our implementation.

Handling the normal vector gradient $\frac{\partial\mathcal{L}}{\partial N}$ is a challenge in training. As the loss function is purely color-based, our normal gradient ultimately comes from the environment map: $\frac{\partial\mathcal{L}}{\partial N}=\frac{\partial\mathcal{L}}{\partial E} \frac{\partial E}{\partial N}$. With $E$ being a texture query, the only non-zero component of $\frac{\partial E}{\partial N}$ comes from the bilinear texture filter. Intuitively, this can be interpreted as the gradient descent process rotating the reflection direction towards the environment map texel best matching the expected pixel color, by updating the underlying $N$. However, the rotation target is restricted to the four texels participating in the bilinear texture filter of a particular query, limiting meaningful gradients to normal vectors already close to the correct value.

Fortunately, our deferred shading model provides an elegant solution. Since we perform the environment lookup at the pixel level, each Gaussian only needs to cover a few pixels with near-correct normal to receive meaningful gradients. We leverage this property to propagate correct normals across neighboring Gaussians, eventually expanding to all reflective surfaces we can find. The details will be explained in the next subsection.

\subsection{Training}

We bootstrap our training process with a short view-independent stage by turning off view-dependent color and reflection optimization, where reflection strengths $r_i$ are initialized to 0 and the per-Gaussian SH color functions $c_i(\mathbf{v})$ are restricted to order 0, i.e., to constant terms $c_i(\mathbf{v})=c_{i,0}$.  With $r_i=0$, reflection-related optimizations are disabled as relevant gradients have zero magnitude. This stage lasts for a few thousand iterations, typically taking a few minutes on a high-end GPU. The optimization process is the same as in the original 3DGS~\cite{3DGS}.

\begin{figure}[tb]
\centering
\includegraphics[width=1.0\columnwidth]{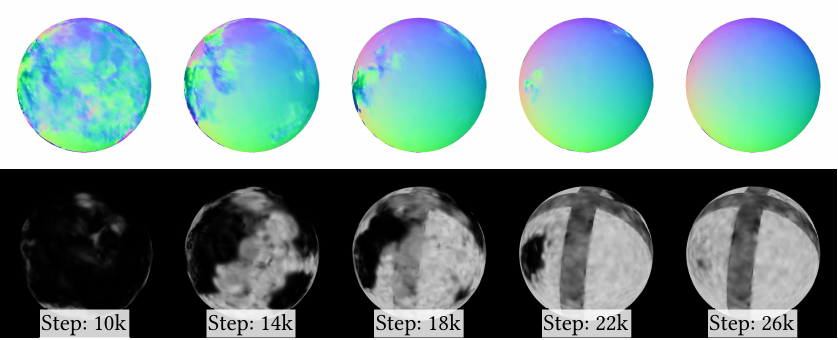}
\caption{The propagation of Gaussian normal and reflection strength at various training steps. \label{fig:surfprop}}
\Description{The first row of the figure is Gaussian normal and the second row is reflection strength of a ball. From left to right are the results of training steps 10k, 14k, 18k, 22k, and 26k. It can be observed that the defects in normals decrease gradually, leading to a perfect sphere.}
\end{figure}

In the following training, we turn on the optimization of per-Gaussian reflection strength $r_i$ and environment map. A few Gaussians may get relatively large reflection strength during optimization (i.e., $r_i>0.1$). Observing that such reflective Gaussians have near-correct normal vectors (see \figref{fig:surfprop}), we propose to
propagate their normal vectors to nearby Gaussians. Specifically, when one Gaussian with near-correct normal overlaps a different Gaussian without one, some shared pixels can also have near-correct normal, which will get meaningful normal gradients and help to optimize the normal of the later Gaussian. To facilitate such propagation, we periodically raise the opacity of all Gaussians to at least 0.9 and the reflection strength to at least 0.001, then scale up the two longest axises of reflective Gaussians by $1.5\times$, leaving the shortest used-as-normal axises intact. This makes almost every reflective Gaussian overlap with its neighbors, and the universally-high opacity ensures that every visible Gaussian contributes a significant magnitude to surface normal and in reciprocal, getting influenced by meaningfully normal gradients during back propagation. We call this process \emph{normal propagation}. 

\figref{fig:surfprop} shows a concrete example of how correct normals propagate as training proceeds. Starting from noisy shortest axis directions generated by constant color fitting, some randomly-correct spots first gained reflection strength by step 9000. They then propagate their correct normal vectors to neighboring Gaussians, gradually spreading over the noisy bumps, and finally yielding a smooth sphere. The reflection strength map increases early but only starts approaching the correct value as normal vectors become more accurate.

The diffuse colors $c_{i,0}$ have already been optimized by the view-independent bootstrap and during reflection training they tend to over-fit, which can hinder reflective surface discovery. To counter this effect, we intentionally sabotage the colors of not-yet-reflective Gaussians with $r_i \le 0.1$, by adding a $\pm10\%$ noise whenever normal propagation is applied. We call this process \emph{color sabotage}.

Our periodic opacity raise conflicts with the periodic opacity clamping in original 3D Gaussian Splatting, which clamps opacity to no more than 0.01 \cite{3DGS}. The color sabotage also prevents the color terms from converging. As a workaround, we interleave opacity-raising periods with opacity-clamping periods so that they are never applied simultaneously. We also terminate normal propagation and color sabotage once the number of Gaussians with $r_i>0.1$ stop increasing for a fixed number of iterations, which indicates that no more specular surfaces can be found. We only start to optimize higher-order color SH coefficients after this specular termination criteria has been met, to prevent them from interfering with reflection training.

\section{Results and Evaluation}

\begin{table*}
\caption{Per-scene image quality comparison on synthesized test views.\label{tab:metrics}}
\tabcolsep=0.10cm
\renewcommand\arraystretch{1.2}
\resizebox{\linewidth}{!}{
\begin{tabular}{cc|cccccc|cccccc|ccc}
\hline
\multicolumn{2}{c|}{}                                         & \multicolumn{6}{c|}{Shiny Blender~\cite{ref_nerf}}                                                                                                                                                                                                                                                                                                    & \multicolumn{6}{c|}{Glossy Synthetic~\cite{nero}}                                                                                                                                                                                                                                                                                                   & \multicolumn{3}{c}{Real}                                                                                                                                           \\ \cline{3-17} 
\multicolumn{2}{c|}{\multirow{-2}{*}{Datasets}}               & ball                                                 & car                                                  & coffee                                               & helmet                                               & teapot                                               & toaster                                              & bell                                                 & cat                                                  & luyu                                                 & potion                                               & tbell                                                & teapot                                               & garden                                               & sedan                                                & toycar                                               \\ \hline
\multicolumn{1}{c|}{}                        & Ref-NeRF       & 33.16                                                & \cellcolor{c1} 30.44 & \cellcolor{c2}33.99                                                & 29.94                                                & 45.12                                                & 26.12                                                & 30.02                                                & 29.76                                                & 25.42                                                & 30.11                                                & 26.91                                                & 22.77                                                & \cellcolor{c1} 22.01 & 25.21                                                & 23.65                                                \\
\multicolumn{1}{c|}{}                        & NPC            & 23.76                                                & 24.19                                                & 30.39                                                & 25.59                                                & 41.22                                                & 19.76                                                & 22.41                                                & 25.35                                                & 23.68                                                & 23.09                                                & 19.03                                                & 18.21                                                & 21.01                                                & 24.77                                                & 22.84                                                \\
\multicolumn{1}{c|}{}                        & 3DGS           & 27.65                                                & 27.26                                                & 32.30                                                 & 28.22                                                & 45.71                                                & 20.99                                                & 25.11                                                & 31.36                                                & 26.97                                                & 30.16                                                & 23.88                                                & 21.51                                                & 21.75                                                & 26.03                                                & \cellcolor{c2}23.78                                                \\
\multicolumn{1}{c|}{}                        & GShader        & 30.99                                                & 27.96                                                & 32.39                                                & 28.32                                                & \cellcolor{c2}45.86                                                & \cellcolor{c2}26.28                                                & 28.07                                                & 31.81                                                & 27.18                                                & 30.09                                                & 24.48                                                & 23.58                                                & 21.74                                                & 24.89                                                & 23.76                                                \\
\multicolumn{1}{c|}{}    & ENVIDR        & \cellcolor{c1}41.02                                                & 27.81                                                & 30.57                                                & \cellcolor{c1}32.71                                                & 42.62                                                & 26.03                                                & \cellcolor{c2}30.88                                                & 31.04                                                & \cellcolor{c2}28.03                                                & \cellcolor{c2}32.11                                                & \cellcolor{c2}28.64                                                & \cellcolor{c1}26.77                                                & 21.47                                                & 24.61                                                & 22.92                                                \\
\multicolumn{1}{c|}{}                     & Ours, forward  & 27.64                                                & 28.99                                                & 31.61                                                & 28.01                                                & 45.68                                                & 24.83                                                & 25.74                                                & \cellcolor{c2}32.22                                                & 27.01                                                & 30.25                                                & 24.11                                                & 23.13                                                & 21.49                                                & \cellcolor{c2}26.05                                                & 23.49                                                \\
\multicolumn{1}{c|}{\multirow{-6}{*}{PSNR $\uparrow$}}  & Ours, deferred & \cellcolor{c2} 33.66 & \cellcolor{c2}30.39                                                & \cellcolor{c1} 34.65 & \cellcolor{c2} 31.69 & \cellcolor{c1} 47.12 & \cellcolor{c1} 27.02 & \cellcolor{c1} 31.65 & \cellcolor{c1} 33.86 & \cellcolor{c1} 28.71 & \cellcolor{c1} 32.29 & \cellcolor{c1} 28.94 & \cellcolor{c2} 25.36 & \cellcolor{c2}21.82                                                & \cellcolor{c1} 26.32 & \cellcolor{c1} 23.83 \\ \hline
\multicolumn{1}{c|}{}                        & Ref-NeRF       & 0.971                                                & \cellcolor{c2}0.950                                                 & \cellcolor{c2}0.972                                                & 0.954                                                & 0.995                                                & 0.921                                                & 0.941                                                & 0.944                                                & 0.901                                                & 0.933                                                & \cellcolor{c2}0.947                                                & 0.897                                                & \cellcolor{c1} 0.584 & 0.720                                                 & 0.633                                                \\
\multicolumn{1}{c|}{}                        & NPC            & 0.908                                                & 0.898                                                & 0.955                                                & 0.938                                                & 0.994                                                & 0.835                                                & 0.892                                                 & 0.921                                                & 0.854                                                & 0.877                                                & 0.742                                                & 0.762                                                & 0.558                                                & 0.711                                                & 0.547                                                \\
\multicolumn{1}{c|}{}                        & 3DGS           & 0.937                                                & 0.931                                                & \cellcolor{c2}0.972                                                & 0.951                                                & \cellcolor{c2}0.996                                                & 0.894                                                & 0.908                                                & 0.959                                                & 0.916                                                & 0.938                                                & 0.900                                                  & 0.881                                                & 0.571                                                & \cellcolor{c2}0.771                                                & \cellcolor{c2}0.637                                                \\
\multicolumn{1}{c|}{}                        & GShader        & 0.966                                                & 0.932                                                & 0.971                                                & 0.951                                                & \cellcolor{c2}0.996                                                & \cellcolor{c2}0.929                                                & 0.919                                                & 0.961                                                & 0.914                                                & 0.936                                                & 0.898                                                & 0.901                                                & 0.576                                                & 0.728                                                & \cellcolor{c2}0.637                                                \\
\multicolumn{1}{c|}{}                        & ENVIDR        &\cellcolor{c1}0.997	&0.943	&0.962	&\cellcolor{c1}0.987	&0.995	&0.922	&\cellcolor{c2}0.954	&\cellcolor{c2}0.965	&\cellcolor{c2}0.931	&\cellcolor{c1}0.960	&\cellcolor{c2}0.947	&\cellcolor{c1}0.957	&0.561	&0.707	&0.549
                                                \\
\multicolumn{1}{c|}{}                        & Ours, forward  & 0.939                                                & 0.941                                                & 0.968                                                & 0.947                                                & \cellcolor{c2}0.996                                                & 0.919                                                & 0.909                                                & 0.964                                                & 0.911                                                & 0.938                                                & 0.904                                                & 0.891                                                & 0.566                                                & 0.767                                                & 0.626                                                \\
\multicolumn{1}{c|}{\multirow{-6}{*}{SSIM $\uparrow$}}  & Ours, deferred & \cellcolor{c2} 0.979 & \cellcolor{c1} 0.962 & \cellcolor{c1} 0.976 & \cellcolor{c2} 0.971 & \cellcolor{c1} 0.997 & \cellcolor{c1} 0.943 & \cellcolor{c1} 0.962 & \cellcolor{c1} 0.976 & \cellcolor{c1} 0.936 & \cellcolor{c2} 0.957 & \cellcolor{c1} 0.952 & \cellcolor{c2} 0.936 & \cellcolor{c2}0.581                                                & \cellcolor{c1} 0.773 & \cellcolor{c1} 0.639 \\ \hline
\multicolumn{1}{c|}{}                        & Ref-NeRF       & 0.166                                                & 0.050                                                 & 0.082                                                & 0.086                                                & 0.012                                                & 0.083                                                & 0.102                                                & 0.104                                                & 0.098                                                & 0.084                                                & 0.114                                                & 0.098                                                & 0.251                                                & 0.234                                                & \cellcolor{c1} 0.231 \\
\multicolumn{1}{c|}{}                        & NPC            & 0.237                                                & 0.120                                                 & 0.119                                                & 0.156                                                & 0.013                                                & 0.226                                                & 0.203                                                & 0.121                                                & 0.101                                                & 0.174                                                & 0.243                                                & 0.246                                                & 0.302                                                & 0.311 & 0.347                                                \\
\multicolumn{1}{c|}{}                        & 3DGS           & 0.162                                                & 0.047                                                & 0.079                                                & 0.081                                                & 0.008                                                & 0.125                                                & 0.104                                                & 0.062                                                & 0.064                                                & 0.093                                                & 0.125                                                & 0.102                                                & \cellcolor{c2}0.248                                                & \cellcolor{c1} 0.206 & \cellcolor{c2}0.237                                                \\
\multicolumn{1}{c|}{}                        & GShader        & 0.121                                                & \cellcolor{c2}0.044                                                & \cellcolor{c2}0.078                                                & 0.074                                                & \cellcolor{c2}0.007                                                & \cellcolor{c1} 0.079 & 0.098                                                & 0.056                                                & 0.064                                                & 0.088                                                & 0.122                                                & 0.091                                                & 0.274                                                & 0.259                                                & 0.239                                                \\
\multicolumn{1}{c|}{}                        & ENVIDR        &\cellcolor{c1}0.020	&0.046	&0.083	&\cellcolor{c1}0.036	&0.009	&\cellcolor{c2}0.081	&\cellcolor{c2}0.054	&\cellcolor{c2}0.049	&\cellcolor{c2}0.059	&\cellcolor{c1}0.072	&\cellcolor{c2}0.069	&\cellcolor{c1}0.041	&0.263	&0.387	&0.345
                                                \\
\multicolumn{1}{c|}{}                        & Ours, forward  & 0.156                                                & \cellcolor{c2}0.044                                                & 0.081                                                & 0.082                                                & 0.008                                                & 0.091                                                & 0.104                                                & 0.059                                                & 0.068                                                & 0.096                                                & 0.124                                                & 0.096                                                & 0.252                                                & 0.221                                                & 0.249                                                \\
\multicolumn{1}{c|}{\multirow{-6}{*}{LPIPS $\downarrow$}} & Ours, deferred & \cellcolor{c2} 0.098 & \cellcolor{c1} 0.033 & \cellcolor{c1} 0.076 & \cellcolor{c2} 0.049 & \cellcolor{c1} 0.005 & \cellcolor{c2}0.081                                                & \cellcolor{c1} 0.046 & \cellcolor{c1} 0.040  & \cellcolor{c1} 0.053 & \cellcolor{c2} 0.075 & \cellcolor{c1} 0.067 & \cellcolor{c2} 0.067 & \cellcolor{c1} 0.247 & \cellcolor{c2}0.208                                                & \cellcolor{c1} 0.231 \\ \hline
\end{tabular}
}
\end{table*}
We conduct comprehensive experiments on a workstation with an i7-13700KF CPU, 32GB memory and an NVIDIA RTX 4090 GPU, to demonstrate the effectiveness and efficiency of our approach. As an ablation study for our core deferred reflection design, we also implement a forward-shading alternative, which computes reflection colors on individual Gaussians and splats them to the final image. The alternative forward renderer is used for both training and testing and other training designs remain unchanged.

\paragraph{Dataset.} We conduct evaluation on several datasets with specular objects, including two synthetic datasets: Shiny Blender~\cite{ref_nerf} and Glossy Synthetic~\cite{nero}, and one real captured dataset used by Ref-NeRF~\cite{ref_nerf}. We also use the non-specular NeRF Synthetic~\cite{nerf} dataset as a regression test.

\paragraph{Implementation details.} For real scenes, we use a spherical domain $M$ to cover the foreground object. We restrict our deferred reflection stage to Gaussians inside $M$ to reduce interference from the background during training. Based on our observations, background objects that are only captured in a limited amount of views exhibit a similar behavior as reflective objects, which interferes with our environment map fitting.

\paragraph{Baselines and metrics.} We compare our method against the following baselines: \textbf{3DGS:} vanilla 3D Gaussian Splatting~\cite{3DGS} with no special treatment for reflection; \textbf{GShader}~\cite{jiang2023gaussianshader}: a method that shades each Gaussian with a reflection-aware shader, which can be considered as a differentiable forward rendering pipeline; \textbf{ENVIDR}~\cite{envidr}: a SDF-based method using neural rendering for inverse rendering; \textbf{Ref-NeRF}~\cite{ref_nerf}: a NeRF-based method focusing on reflective objects rendering; \textbf{NPC:} Neural Point Catacaustics~\cite{catacaustics}: a renderer that warps a point-based hallucinated reflection using MLPs. All methods are applied to the same input data, except for NPC, which requires a mask covering all reflectors for each input image. We use the foreground mask estimated by~\cite{kirillov2023segment} as a substitute. We present quantitative results measured with three standard metrics: PSNR, SSIM and LPIPS. 

To demonstrate the accuracy normal and light reconstruction, we also compare our method with SDF-based inverse rendering methods: \textbf{ENVIDR}~\cite{envidr} and \textbf{NVDiffRec}~\cite{nvidiffrec}. We use Mean Angular Error in degrees (MAE$^\circ$) to evaluate the normal reconstruction accuracy. We evaluate environment map reconstruction accuracy with LPIPS to mitigate the fundamental ambiguity between albedo, roughness and light.

\subsection{Comparisons with baselines}

\begin{figure*}
\includegraphics[width=0.98\linewidth]{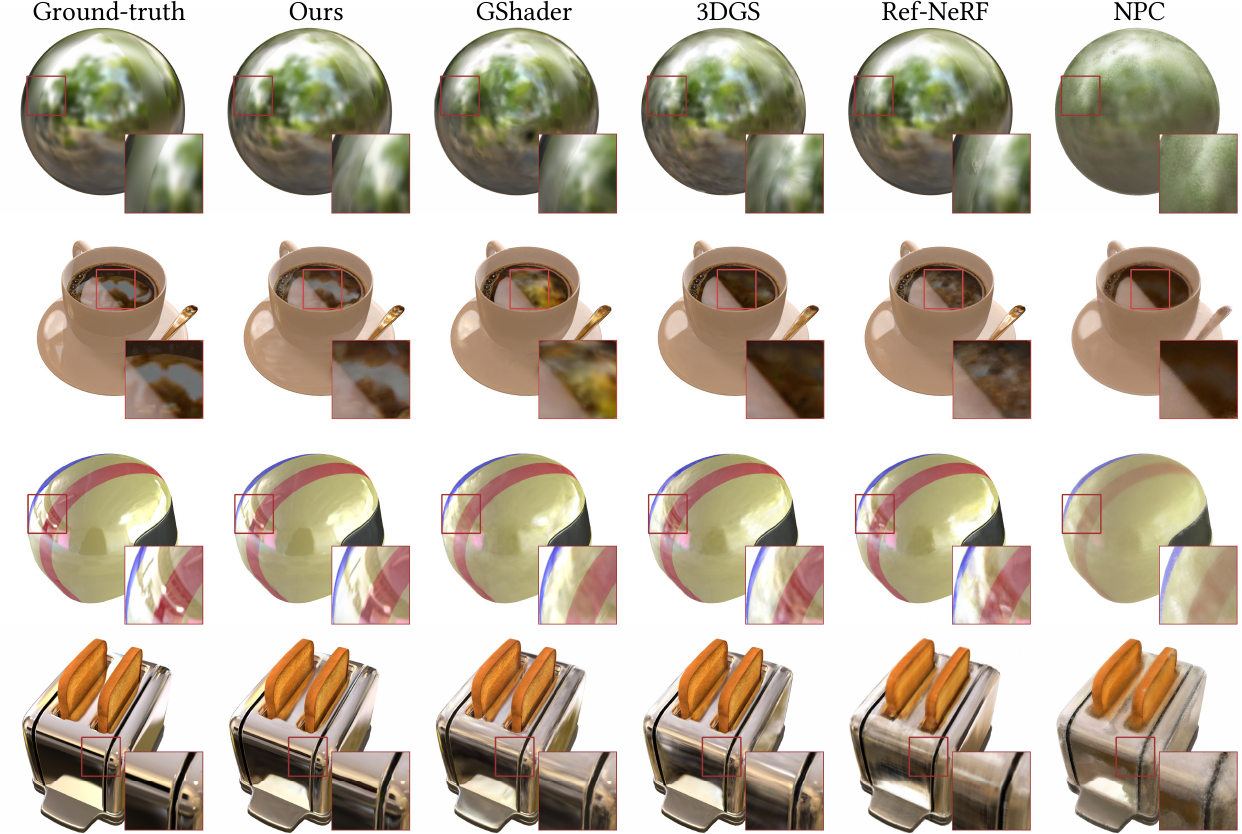}
\caption{Qualitative comparisons on synthetic scenes. From top to bottom: ball, coffee, helmet, toaster.\label{fig:synth}}
\Description{In the figure, our method clearly displays reflections of trees within the sphere, the coffee liquid surface, and the helmet, as well as reflections from the corner of the toaster.}
\end{figure*}

\begin{figure*}
\includegraphics[width=0.98\linewidth]{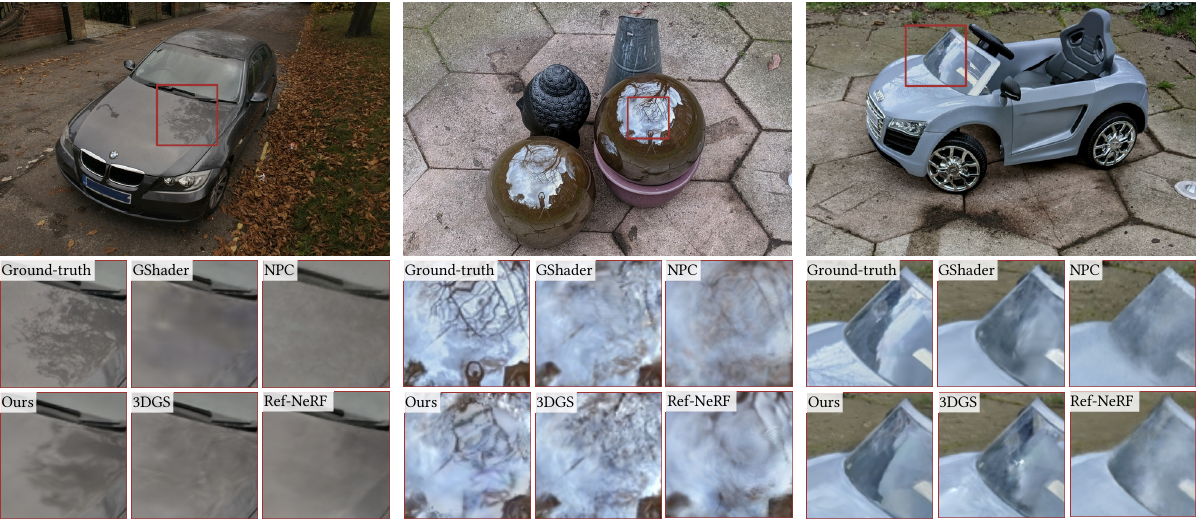}
\caption{Qualitative comparisons on real scenes. From left to right: sedan, garden, toycar.\label{fig:real}}
\Description{From left to right, our method clearly displays reflections of branches on the front hood of the car, reflections of branches on the smooth stone sphere, and reflections of the front windshield glass of the toy car. The reflections in other methods appear more blurred.}
\end{figure*}

\begin{figure*}
\centering
\includegraphics[width=0.95\linewidth]{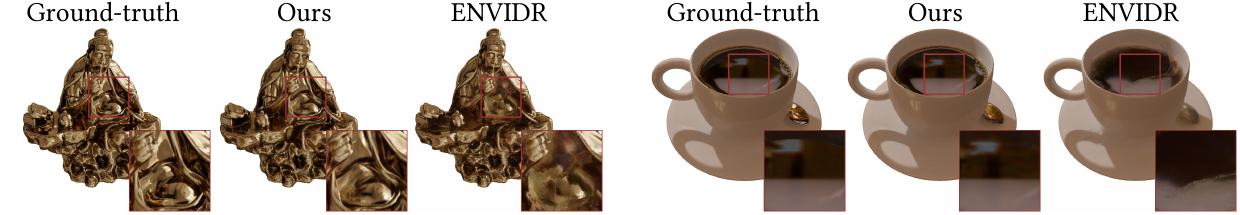}
\caption{Qualitative comparisons between ENVIDR~\cite{envidr} and our method on luyu and coffee synthetic scenes.} \label{fig:envidr}
\Description{On the left of the figure is luyu, a reflective Chinese traditional figurine, and on the right is coffee. In the ENVIDR method's results, the sculpted folds of the clothing are blurry, and there is no reflection of the coffee liquid surface.}
\end{figure*}

\paragraph{Image quality.} \tabref{tab:metrics} presents the quantitative comparison results on three datasets. Our method demonstrates a clear advantage in terms of image quality on synthetic datasets and also shows comparable results on real datasets. The visual comparisons on synthetic and real datasets are shown in \figref{fig:synth}, \figref{fig:real} and \figref{fig:envidr}. Ref-NeRF~\cite{ref_nerf} shows some regular patterns in the toaster scene. The results from NPC~\cite{catacaustics} are noisy and tend to exhibit a diffuse texture. The MLP-predicted warping field is severely strained when the camera viewpoint has a high degree of freedom and the reflector contains both diffuse and specular components. The results from 3DGS~\cite{3DGS} and GaussianShader~\cite{jiang2023gaussianshader} are blurred on reflection surfaces or show incorrect reflections. The results from ENVIDR~\cite{envidr} over-smooth some local details. Our method produces much smoother-looking surfaces without blurring the reflection details, and stays closer to the ground truth. In close-up views, our result exhibits significantly less visible Gaussian boundaries, typically manifesting as oval-shaped color blobs. 

\begin{table}[t]
\caption{Normal and light reconstruction quality (evaluated by MAE$^\circ$ and LPIPS respectively) comparisons on the Shiny Blender Dataset. }\label{tab:recon}
\resizebox{0.8\columnwidth}{!}{
\begin{tabular}{c||cccc}
\hline
      & GShader & NVDiffRec & ENVIDR & Ours  \\ \hline
MAE$^\circ$ $\downarrow$   & 22.31   & 17.02     & \cellcolor{c1}4.618  & \cellcolor{c2}4.871 \\
LPIPS $\downarrow$ & 0.621   & 0.636     & \cellcolor{c2}0.615  & \cellcolor{c1}0.511 \\ \hline
\end{tabular}
}
\end{table}

\begin{table}[t]
\caption{Training time and rendering frame rates. \label{tab:performance}}
\resizebox{0.8\columnwidth}{!}{
\begin{tabular}{c||cc|cc}
\hline
         & \multicolumn{2}{c|}{Shiny Blender} & \multicolumn{2}{c}{Real scenes} \\ 
         & Training time          & FPS          & Training time       & FPS       \\ \hline
Ref-NeRF & 19h                    & 0.06         & 27h                 & 0.02      \\
NPC      & 12h                    & 4            & 23h                 & 1         \\
ENVIDR      & 3.2h                    & 3            & 4.7h                 & 1         \\
3DGS     & \cellcolor{c1}6min                   & \cellcolor{c1}277          & \cellcolor{c1}17min               & \cellcolor{c1}84        \\
GShader  & 60min                     & 51           & 72min                & 31        \\
Ours      & \cellcolor{c2}16min                  & \cellcolor{c2}251          & \cellcolor{c2}47min               & \cellcolor{c2}80        \\ \hline
\end{tabular}
}
\end{table}

\begin{table}[t]
\caption{Number of Gaussians in the ball, car, helmet and toaster scenes of the Shiny Blender Dataset. \label{tab:g_num}}
\resizebox{0.8\columnwidth}{!}{
\begin{tabular}{c||cccc}
\hline
               & ball  & car  & helmet & toaster \\ \hline
Ours, deferred & \cellcolor{c1}27.2k & \cellcolor{c1}117k & \cellcolor{c1}36k    & \cellcolor{c1}100k    \\
Ours, forward  & \cellcolor{c2}76.4k & \cellcolor{c2}205k & \cellcolor{c2}62k    & \cellcolor{c2}236k    \\
GShader        & 76.5k & 316k & 106k   & 336k    \\
3DGS           & 199k  & 307k & 100k   & 342k    \\ \hline
\end{tabular}
}
\end{table}

\begin{figure*}
\centering
\includegraphics[width=0.95\linewidth]{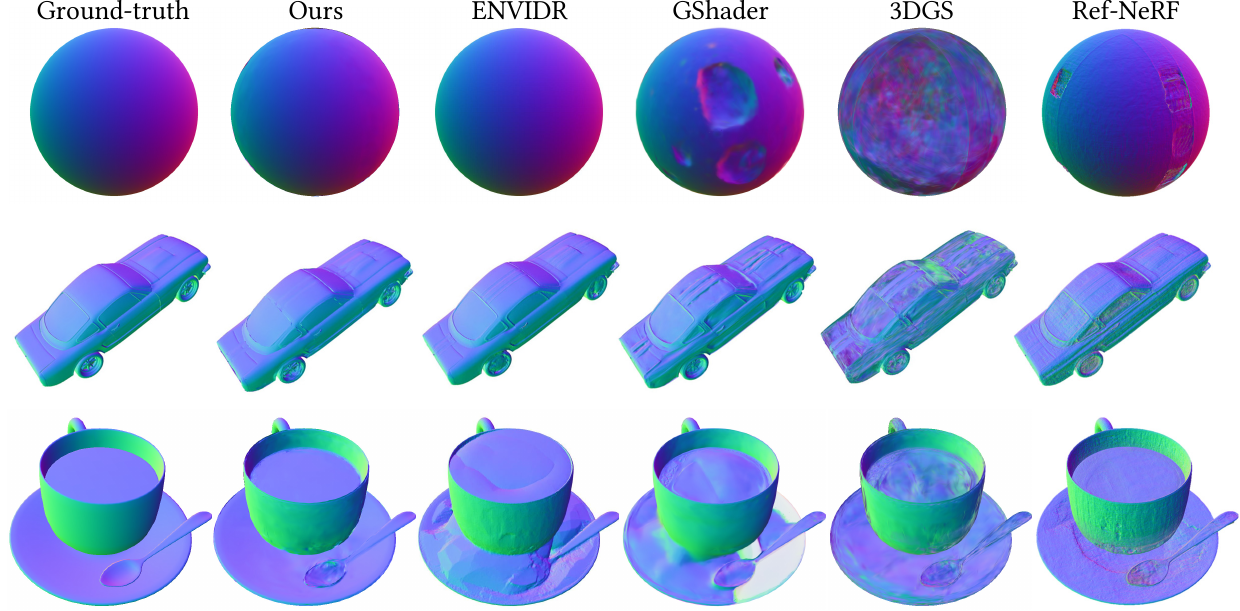}
\caption{Qualitative comparisons of normal produced by different methods. \label{fig:normal}}
\Description{From top to bottom, are respectively the sphere, car, and coffee. Our method exhibits smooth normals close to ground truth. ENVIDR produces incorrect normals above the coffee cup's liquid surface, while GShader and Ref-NeRF displays significant defects on the surface of the sphere.}
\end{figure*}

\paragraph{Normal reconstruction.} Behind the scenes, the quality improvement is backed by improved fitting quality for surface normal and environment map. \tabref{tab:recon} lists the dataset-averaged mean angular error of normal maps on the Shiny Blender Dataset. Despite our method only focusing on the normal of specular surfaces, we still achieve quality comparable to ENVIDR~\cite{envidr}. \figref{fig:normal} compares the normal vectors predicted by our method, ENVIDR, Ref-NeRF~\cite{ref_nerf}, and GaussianShader~\cite{jiang2023gaussianshader}, alongside ground truth and the 3DGS shortest-axis initialization. As illustrated, our method creates smooth results while preserving sharp boundaries from color loss terms alone, like the sharp separation between the saucer and the cup in the coffee scene. ENVIDR, based on the SDF representation, provides nearly perfect normals for the sphere but fails on detailed geometry, such as car wheel rims and thin cup walls, due to the smoothness of SDF. Ref-NeRF produces sharp, yet noisy results. GaussianShader filters out the noise with a smoothness prior, which unfortunately also blurs out object boundaries like between the saucer and the cup in the coffee scene, and shows some surface defects on the ball.

\begin{figure*}\centering
\includegraphics[width=1.0\linewidth]{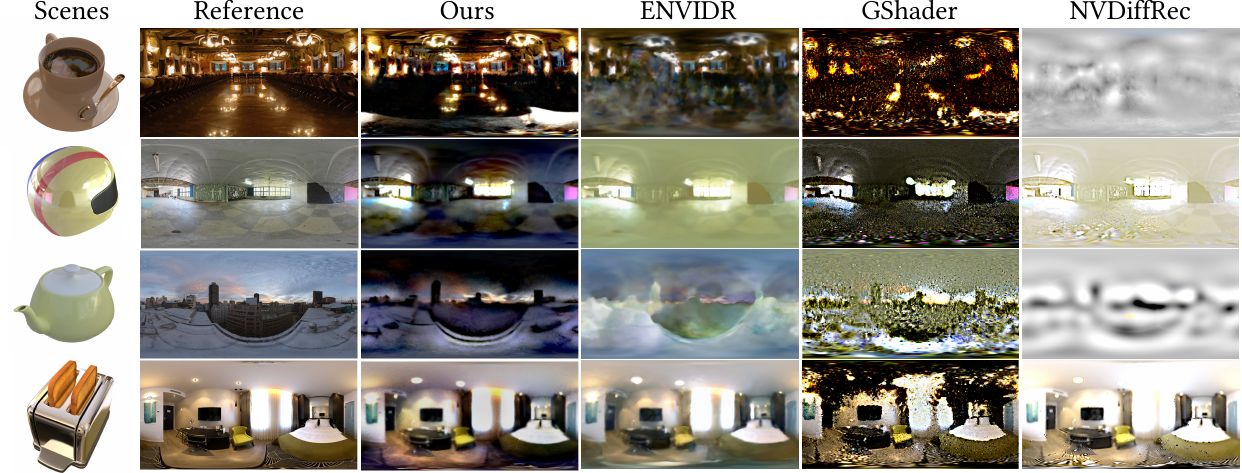}
\caption{Qualitative comparisons of environment maps estimated by different methods.} \label{fig:envmap}
\Description{In the figure, our method has the best quality of environment lighting. ENVIDR presents a foggy appearance in the environment lighting maps of the coffee and teapot, GShader results in significant noise, and NVDiffRec produces completely black-and-white blurred results in the environment lighting maps of the coffee and teapot.}
\end{figure*}

\paragraph{Light reconstruction.} We quantitatively compare the quality of light reconstruction in~\tabref{tab:recon}. We achieve the best quality thanks to our simple rendering model, which reduces the ill-posedness of the inverse problem. We also visualize the environment maps reconstructed by GaussianShader, ENVIDR, NVDiffRec, and our method. To compensate for the fundamental light-albedo ambiguity, we equalize the total energy of the two environment maps before comparison. As illustrated in~\figref{fig:envmap}, the per-pixel reflection allows our method to reconstruct a significantly less noisy environment map with almost full directional coverage, whereas GaussianShader can only reconstruct a few texels for each Gaussian, leading to a noisy image with many holes. For SDF-based methods (ENVIDR and NVDiffRec), they struggle to perfectly decompose lighting, geometry, and material, leading to non-robust lighting estimation.

\begin{figure}[t]
\centering
\includegraphics[width=1.0\columnwidth]{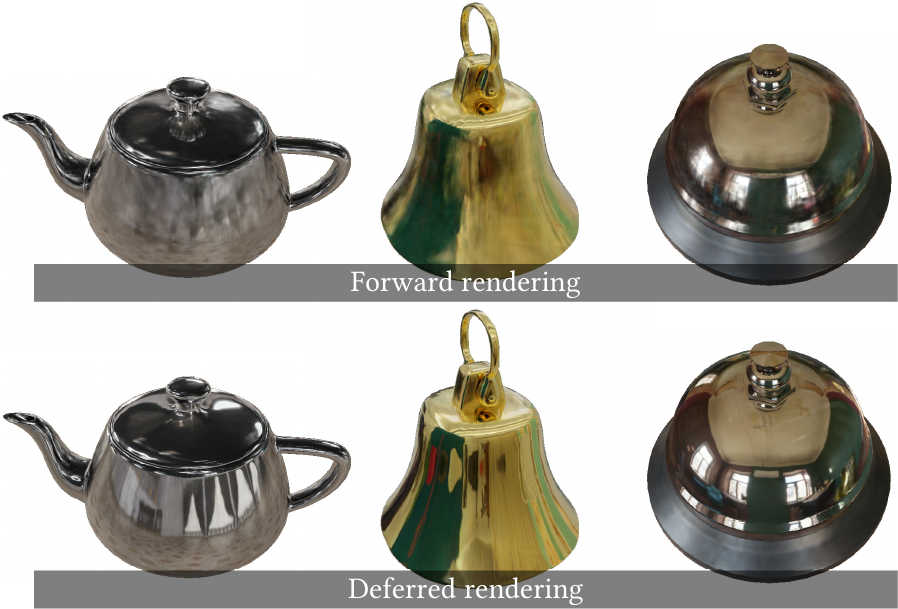}\\
\caption{Quality comparison between forward and deferred designs. From left to right: teapot, bell, tbell. \label{fig:deferred:vs:forward}}
\Description{ From left to right, respectively, are teapot, bell, and tbell. The top row displays the results of forward rendering, while the bottom row shows the results of deferred rendering. The details in forward rendering are mostly blurred.}
\end{figure}

\begin{figure}[t]
\centering
\includegraphics[width=1.0\columnwidth]{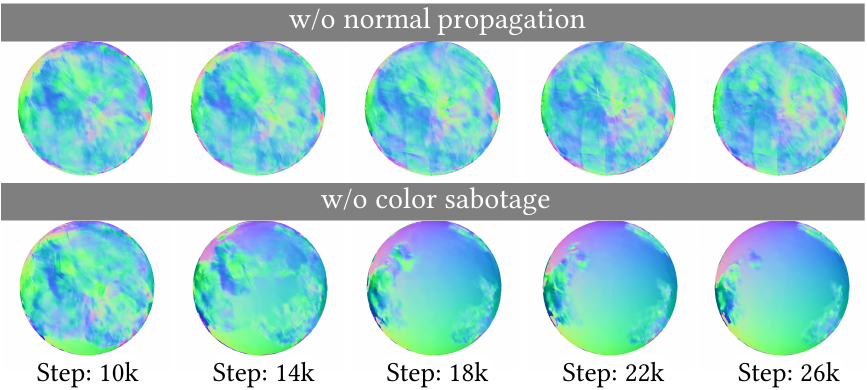}
\caption{Normal maps at various training steps with one algorithm component disabled.}
\label{fig:ablation:disable}
\Description{The top row depicts the results without normal propagation, while the bottom row shows the results without color sabotage. From left to right are the normal results of a sphere at training steps 10k, 14k, 18k, 22k, and 26k. When there is no normal propagation, the normals of the sphere remain chaotic; when there is no color sabotage, there are still some defects in the normals of the sphere at 26k.}
\end{figure}

\paragraph{Efficiency.} \tabref{tab:performance} lists the dataset-averaged training time and rendering frame rate of all tested methods. Frame rate values are computed as the reciprocal of averaged frame render time to reduce the impact of unfairly large numbers on simple scenes. As a reflection-oblivious reference, 3DGS~\cite{3DGS} serves as the performance upper-bound. The NeRF-based method Ref-NeRF~\cite{ref_nerf}, SDF-based method ENVIDR~\cite{envidr}, and point-based method NPC~\cite{catacaustics} require dozens of hours to train. GaussianShader~\cite{jiang2023gaussianshader} requires an hour to train and renders several times slower than vanilla 3DGS. Our method trains to convergence in under an hour and renders at frame rates almost identical to the 3DGS upper bound. We also compare the final number of Gaussians generated by each method in~\tabref{tab:g_num}. By deferring shading to pixels, our method no longer needs to split Gaussians to meet the shading frequency demand. We are able to fit the same scene with significantly less Gaussians, which also improves our real-time performance.

\begin{table}[t]
\caption{The fitting quality impact of disabling various algorithm components. \label{tab:abla}}
\resizebox{0.8\columnwidth}{!}{
\begin{tabular}{c||ccc}
\hline
Ablations       & PSNR $\uparrow$  & SSIM $\uparrow$  & LPIPS $\downarrow$ \\ \hline
Ours            & \cellcolor{c1}33.66 & \cellcolor{c1}0.979 & \cellcolor{c1}0.098 \\
w/o propagation & 27.85 & 0.938 & 0.159 \\
w/o sabotage    & 30.00 & 0.959 & 0.128 \\
\hline
\end{tabular}
}
\end{table}

\begin{table}[t]
\caption{Regression test on non-specular scenes.\label{tab:nonspec}}
\resizebox{0.9\columnwidth}{!}{
\begin{tabular}{ccccccc}
\hline
\multicolumn{7}{c}{NeRF Synthetic}                                                                                                                                                                                                                                                                                                             \\
     & drums                                                & ficus                                                & hotdog                                               & lego                                                 & mic                                                  & ship                                                 \\ \hline
\multicolumn{7}{c}{PSNR $\uparrow$}                                                                                                                                                                                                                                                                                                                       \\ \hline
\small 3DGS & 25.10                                                 & \cellcolor{c1}{\cellcolor{c1} 28.14} & 35.52                                                & \cellcolor{c1}{\cellcolor{c1} 32.94} & 31.55                                                & 29.06                                                \\
Ours & \cellcolor{c1}{\cellcolor{c1} 25.31} & 28.03                                                & \cellcolor{c1}{\cellcolor{c1} 35.58} & \cellcolor{c1}{\cellcolor{c1} 32.94} & \cellcolor{c1}{\cellcolor{c1} 31.97} & \cellcolor{c1}{\cellcolor{c1} 29.07} \\ \hline
\multicolumn{7}{c}{SSIM $\uparrow$}                                                                                                                                                                                                                                                                                                                       \\ \hline
\small 3DGS & \cellcolor{c1}{\cellcolor{c1} 0.947} & \cellcolor{c1}{\cellcolor{c1} 0.965} & \cellcolor{c1}{\cellcolor{c1} 0.983} & \cellcolor{c1}{\cellcolor{c1} 0.979} & 0.986                                                & \cellcolor{c1}{\cellcolor{c1} 0.897} \\
Ours & 0.946                                                & 0.963                                                & 0.982                                                & 0.978                                                & \cellcolor{c1}{\cellcolor{c1} 0.987} & 0.894                                                \\ \hline
\multicolumn{7}{c}{LPIPS $\downarrow$}                                                                                                                                                                                                                                                                                                                      \\ \hline
\small 3DGS & \cellcolor{c1}{\cellcolor{c1} 0.055} & 0.540                                                 & \cellcolor{c1} 0.032          & \cellcolor{c1}{\cellcolor{c1} 0.025} & \cellcolor{c1}{\cellcolor{c1} 0.028} & \cellcolor{c1}{\cellcolor{c1} 0.124} \\
Ours & \cellcolor{c1}{\cellcolor{c1} 0.055} & \cellcolor{c1}{\cellcolor{c1} 0.055} &  0.033 & 0.026                                                & \cellcolor{c1}{\cellcolor{c1} 0.028} & 0.129                                                \\ \hline
\end{tabular}
}
\end{table}

\subsection{Ablation Study} We conduct various ablation studies to validate the impact of key design choices. Specifically, we compare deferred with forward shading, assess the necessity of normal propagation and color sabotage. We also perform a regression test on the non-specular NeRF Synthetic Dataset~\cite{nerf}. 

\paragraph{Deferred shading versus forward shading.} As shown in \tabref{tab:metrics}, using deferred shading in our method results in better PSNR, SSIM and LPIPS, compared to forward shading in all scenes. As illustrated in \figref{fig:deferred:vs:forward}, deferred shading produces sharp reflections with pixel-level details like the window frames in the rightmost column. The leftmost teapot scene also demonstrates a more complete reconstruction of the reflected environment, producing precise curtain shapes while the forward pipeline fails to capture. We summarize the merits of deferred shading for inverse rendering as two-fold:

\begin{itemize}
\item The pixel-precise normal maps alleviate the impact of imprecise Gaussian normal, while also allowing Gaussians with correct normal to propagate gradients to incorrect ones.
\item The environment map is more closely linked to the image loss, which accelerates its optimization and in turn helps guiding normal vectors towards the correct direction.
\end{itemize}

\paragraph{Necessity of normal propagation and color sabotage.} With the respective algorithm component disabled, \figref{fig:ablation:disable} shows the normal maps at various training steps. \tabref{tab:abla} evaluates the quantitative image metrics after convergence. Without normal propagation, the normal map stays almost unchanged for the entire training. Without color sabotage, normal propagates significantly slower and converges prematurely, outpaced by overfit Gaussian colors. Both result in a significant drop of final image quality.

\begin{figure}\centering
\includegraphics[width=1.0\linewidth]{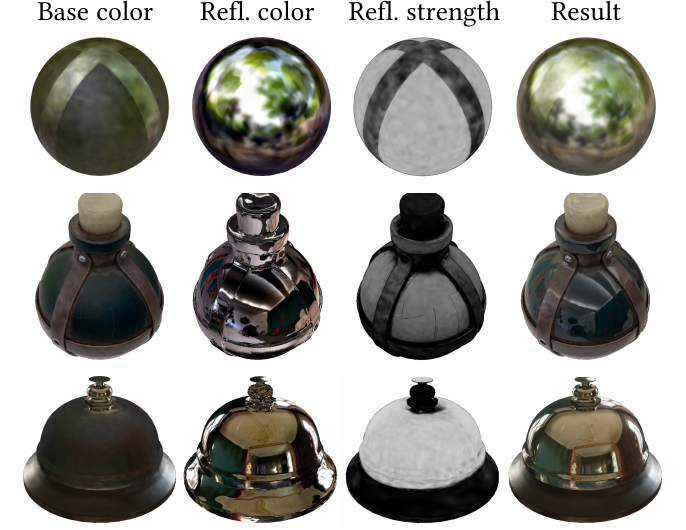}
\caption{Decomposition results of our method. From top to bottom: ball, potion, tbell.} \label{fig:decomp}
\Description{From left to right are the base color, reflection color, reflection strength, and final result images of the ball, potion, and tbell. In the base color map, rough reflections on the strips of the ball and tray of the tbell are presented, while other high-frequency reflections are presented in the reflection color map. The reflection intensity map correctly combines both to form the final result.}
\end{figure}

\paragraph{Decomposition results.} We visualize the base color map, reflection color map, reflection strength map and final results in~\figref{fig:decomp}. Our method precisely captures the specular reflections while leaving other effects to base SH colors, like the rough bands in the sphere scene and the glossy rim of the tbell scene.

\paragraph{Regression in non-specular scene.} \tabref{tab:nonspec} compares our method with plain 3DGS~\cite{3DGS} on non-specular scenes. We are able to achieve an almost indistinguishable quality with minimal regression. As we use the same image loss function with no extra regularization, we are able to correctly estimate the reflection strength as zero for diffuse objects. While our normal propagation depends on reflection, the SH shading we retain does not use normal at all and normal quality on diffuse objects is inconsequential.

\begin{figure}
\centering
\includegraphics[width=1.0\columnwidth]{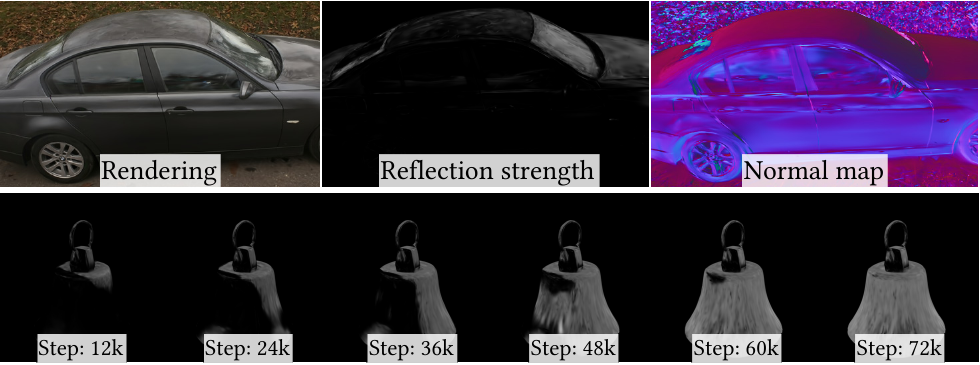}
\caption{Limitations. Top row: inconsistent treatment of car windows. Bottom row: slow convergence on concave bell.}
\label{fig:limit}
\Description{In the top row, the reflection strength of the side car window is 0 and the its normals are chaotic. In the bottom row, normal propagation for the bell is not complete until the 72k training step.}
\end{figure}

\subsection{Limitations}

Our method can handle at most one layer of reflective materials per pixel, which is inherited from traditional deferred shading. The car windows in the top row of~\figref{fig:limit} show two different behaviors of our algorithm on transparent objects. The front window converges to a purely reflective surface and the side windows converges to purely transparent surfaces. A fundamental solution would first require a way to reliably create Gaussians on the transparent windows during initialization.

Normal propagation works less efficient on concave scenes, like the bell scene in the bottom row of~\figref{fig:limit}. Training still converges but takes considerably more time. A better designed optimization strategy can potentially fix this issue.

\section{Conclusion}

We have presented a high-quality deferred Gaussian splatting renderer specializing in reflection. It demonstrates stable training and competitive visual quality at almost identical frame rates to vanilla 3D Gaussian splatting, also producing accurate surface normal and environment maps. 

Our deferred shading approach may open up many possibilities for future exploration.
It would be interesting to explore more creative splits of the rendering equation in the context of Gaussian splatting.
Our pipeline can also be extended to higher quality reflection algorithms beyond an environment map, including screen-space reflections~\cite{mcguire2014efficient} and hardware ray tracing. Generalizing 3D Gaussians and differentiable rendering to such methods can lead to significantly better reflection qualities. It is also interesting to explore the possibility of adding a physically-based roughness, generalizing our method to glossy materials.

\begin{acks}
This work is partially supported by NSF China (No. 62227806 \&
U23A20311) and the XPLORER PRIZE. The source code and data are available at \url{https://github.com/gapszju/3DGS-DR}.
\end{acks}

\bibliographystyle{ACM-Reference-Format}
\bibliography{dr3dgs}

\end{document}